\documentclass[runningheads]{llncs}
\usepackage{graphicx}

\begin{document}
\title{Uncovering Gender Bias in Media Coverage of Politicians with Machine Learning}

\author{Susan Leavy}
\authorrunning{S. Leavy}

\institute{University College Dublin, Ireland \\
\email{susan.leavy@ucd.ie}}
\maketitle              
\begin{abstract}

This paper presents research uncovering systematic gender bias in the representation of political leaders in the media, using artificial intelligence. Newspaper coverage of Irish ministers over a fifteen year period  was gathered and analysed with natural language processing techniques and machine learning. Findings demonstrate evidence of gender bias in the portrayal of female politicians, the kind of policies they were associated with and how they were evaluated in terms of their performance as political leaders. This paper also sets out a methodology whereby media content may be analysed on a large scale utilising techniques from artificial intelligence within a theoretical framework founded in gender theory and feminist linguistics. 

\end{abstract}
\section{Introduction}

The prevalence of gender bias in political news coverage has been consistently highlighted by both academics and politicians (Miller and Peake, 2013; Norris, 1997; Trimble et al., 2013). Such bias has been shown to deter women from entering political life (Heldman et al., 2005; Fox and Lawless, 2004).  Given issues concerning the global imbalance of women in politics (IPU, 2013) it is important that systematic ways of identifying gender bias in the media are developed. 
In studies of the representation of women in the media content analysis of text is the method predominantly used.  Given the time consuming nature of this approach, the volume of text studied can be limited (Baker 2014; Neuendorf, 2011).  This paper explores the use of new automated techniques, employing natural language processing and machine learning to analyse a large corpus of newspaper text. Text classification has been used in similar ways to examine ideological differences expressed in political speeches (Diermeier et al., 2012; Yu et al., 2008) and gender differences among authors (Boulis and Ostendorf, 2005; Corney et al., 2002; Otterbacher, 2010; S. Argamon and Shimoni, 2003). While these studies are primarily author focused, this research will build on these methodologies to analyse the language of journalists as a collective. In this sense, for the purposes of building on existing methodologies, the author is considered analogous to the media organisation and systematic gender bias is attributed to it rather than to individual journalists.  
The newspaper articles analysed included 15 years of newspaper coverage of ministerial level Irish politicians between 1997 and 2011.  This date range was selected as it represented a period in Irish political history where the same political party remained the largest party in government and the political figures holding office remained largely unchanged. This consistency in political power presents an opportunity to study the representation of politicians over a long period of time thus mitigating the influence of particular events or personalities on the language in the media.
This study focused on two newspapers in Ireland with the largest circulation, The Irish Times and The Irish Independent (Table \ref{tbl_1}). Newspaper content was selected rather than digital sources, as this is the primary source of text-based news in Ireland. The Irish Times is traditionally considered a paper of record and therefore could be a powerful indicator of the prevalence of gender bias in the media. With the aim of facilitating the broadening of the scope of this research in the future, the methodology outlined in this research is designed with scalability in mind so that it could be applied to text from a different media source.
\begin{table}[ht]
\centering
\begin{tabular}{c}
\includegraphics[width=0.66\textwidth]{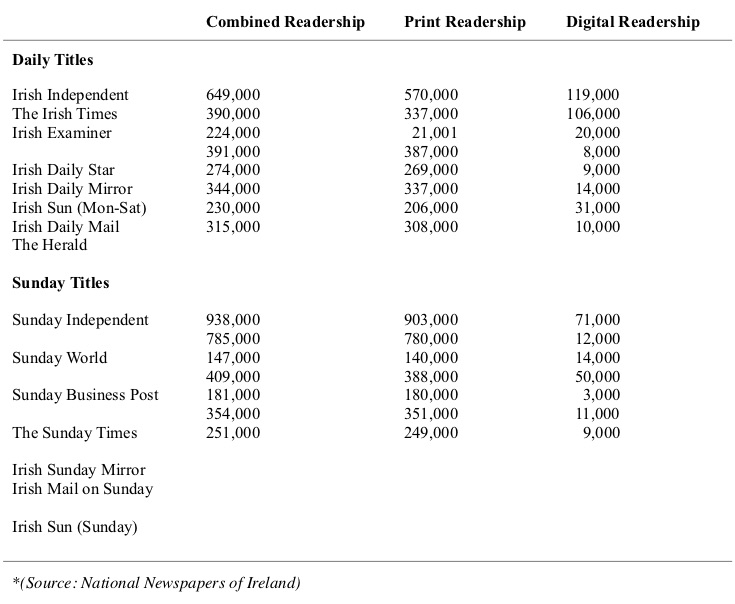}
\end{tabular}
\caption{Irish Newspaper Readership.} \label{tbl_1}
\end{table}

A range of natural language processing techniques were applied to extract information from the newspaper text and machine learning algorithms were then used to build models to automatically classify articles according to whether they featured male or female politicians. The accuracy in the predictive models was indicative of the extent of differences in coverage of male and female politicians. These models were then analysed to assess whether differences in the coverage were attributable to gender bias or not. The context that the gender differences occurred was examined using concordance analysis.

\section{Related Research}

Studies have shown how systematic gender bias can be embedded in how language is used. For example, bias has been identified in ways men and women are described (Mills, 2002; Sigley and Holmes, 2002; Baker, 2010). In portrayals of female politicians references to their political style tends to be more negative and invoke stereotypes more often (Trimble et al., 2013; Pantti, 2005; Gidengil and Everitt, 2003; Ross, 2000). Naming conventions used to refer to politicians often attribute higher status to male politicians (Uscinski and Goren, 2011).  

The kind of coverage that female politicians receive can also indicate gender bias. Studies have shown that female politicians feature less in the media than their male counterparts (Bystrom et al., 2001; Miller et al., 2010; Pearce, 2008). The coverage they do receive tends to focus more on their personal life and family (Miller and Peake, 2013; Trimble et al., 2013; Garcia-Blanco and Wahl-Jorgensen, 2011; Miller et al., 2010). Male politicians are more likely to be associated with policy issues regardless of their political output or role and these associations are more likely to conform to gender stereotypes (Kahn, 1996; Kahn and Goldenberg, 1991; Scharrer, 2002). This research builds on the findings of these studies and explores whether using automated methods could facilitate large-scale analysis of corpora to systematically identify gender bias in the media.  

Text classification studies have addressed a range of issues such as an author’s gender, political ideology, subjectivity and an author’s personality type. While these are not directly analogous to the topic of gender bias, they have uncovered ways in which information regarding subjective opinion or personal attributes can be encoded in language. The advantage of using machine learning algorithms to analyse text is the potential it has to uncover unexpected patterns. This can be achieved by allowing a machine learning algorithm analyse all words in a corpus, minimising the amount of encoding or pre-processing applied to the text (Argamon et al., 2009; Boulis and Ostendorf, 2005; Nowson and Oberlander, 2006; Opsomer et al., 2008; Kucukyilmaz et al., 2006; Otterbacher  2010).  Koppel et al. (2009) achieved almost perfect accuracy in classifying religious texts according to their source organisation or ideological stance simply by using 1,000 most common words in a corpus.  Other studies have extracted only those attributes of text that indicate writing style and tested their correlation with gender by evaluating the classification accuracies of a machine-learning algorithm (Corney et al., 2002; Hota et al., 2006; Koppel et al., 2002; Sabin et al., 2008). Diermeier et al. (2012) used text classification to explore differences in speeches by liberal and conservative politicians in the US Senate using words and corresponding syntactic information as features. Yu et al. (2008) in a similar study examined a collection of US Senate and House speeches using words as features, excluding rare and common words. 

Where machine learning is used to analyse language for specific patterns in a data-driven approach, feature engineering and text pre-processing founded in a theoretical approach to the classification topic plays a significant role. Pang and Lee (2008) classified movie reviews into either positive or negative categories based on single word features. However, improved results were found by including the place of words in sentences (Kim and Hovy, 2006) and by pre-processing the text into word sets (Dave. et al., 2003). In sentiment extraction the use of lexicons of sentiment bearing terms for feature extraction has been shown to be effective (Riloff et al., 2006).

Using machine learning to analyse media content has the potential to uncover new language patterns along with analysing the text for specific linguistic features. In applying this approach to the topic of gender bias in media coverage of politicians, this research aimed to examine whether the kind of gender bias identified in previous studies is evident in Irish newspapers and also to facilitate the discovery of previously undiscovered forms of gender bias. There are two research questions for this study, reflecting the dual aim of exploring a new methodology for analysing gender bias and evaluating the prevalence of gender bias in the Irish media. The research questions for this study are as follows: (1) How can text classification be used to uncover differences in the coverage of male and female politicians (2) What differences in the coverage of male and female politicians could be attributed to gender bias.

\section{Method}

A corpus was created comprising newspaper coverage of Irish Politicians between 1997 and 2011. Female cabinet ministers and comparable male ministers were identified (Table \ref{tbl_2}).  Male and female ministers were considered comparable if they held the same portfolio during the 15-year timeframe. Newspaper content featuring the politicians during the time period where they held comparable ministerial portfolios was gathered from Ireland’s leading newspapers, The Irish Times, The Independent and the Sunday Independent.  The corpus contained 47,981 newspaper articles.

\begin{table}[ht]
\centering
\begin{tabular}{c}
\includegraphics[width=0.66\textwidth]{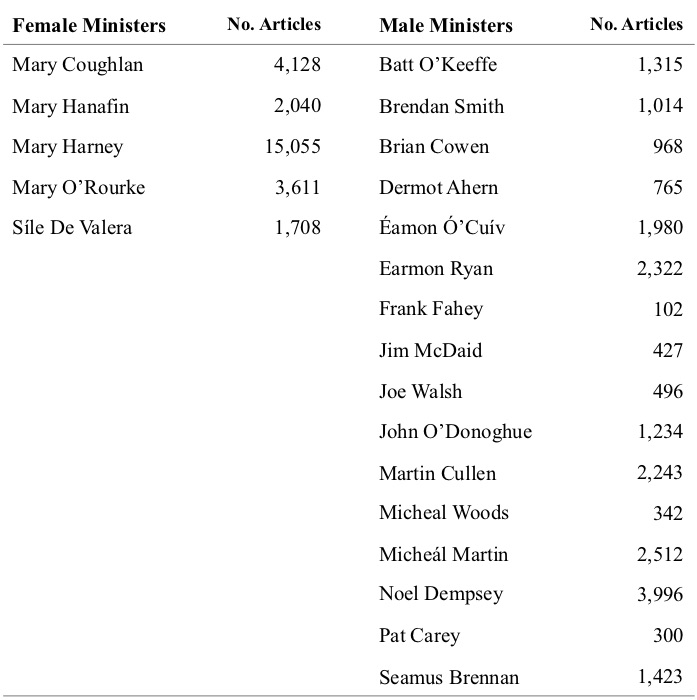}
\end{tabular}
\caption{ Sample of Articles Featuring Irish Cabinet Ministers.} \label{tbl_2}
\end{table}

The articles were labelled according to whether they featured a male or female politician. Where an article featured both male and female politicians, multiple instances of the article were created with each label. The texts were pre-processed to remove attributes such as pronouns that would signal the gender of the politician to a machine-learning algorithm but would have no broader meaning in relation to gender bias. Quotations and paraphrasing of the politicians remained included in the corpus. How quotations are selected for inclusion in newspaper articles has been cited as a gendered process whereby quotations from male and female politicians are treated and presented differently (Kroon 2006). Quotations were therefore included in the corpus in order to highlight any differences in how male and female politicians were being quoted. 	

Features were then extracted from the articles and represented as an array of vectors. Multiple approaches to text representation were explored including a bag-of-words, tf-idf and Boolean representation. The features included linguistic features of the text along with structural features of the articles (Table \ref{tbl_3}).

\begin{table}[ht]
\centering
\begin{tabular}{l}
Feature Sets \\ \hline
Unigrams \\
Adjectives \\
Verbs \\
Semantic Lexicons \\
Newspaper Sections \\
\end{tabular}
\caption{Feature Sets Used in Text Classification.} \label{tbl_3}
\end{table}

The approach to feature selection incorporated both a data-driven and a data-based approach (Baker, 2014). The data-driven approach involved using words as features and keeping text pre-processing to a minimum to allow patterns emerge from the text. Specific parts of speech, namely adjectives and verbs were also used as features to implement a data-based approach and assess the correlation of certain feature types with the gender of the politician. In another data-based approach, the correlation of gender with language associated with power and action was assessed by extracting features based on the General Inquirer lexicon of power and action words (Stone, 1966). In order to identify features that were most closely associated with politicians, linguistic features that occurred in the same sentence as the mention of a politician’s name were extracted. Structural features such as the sections of the newspaper that the articles appeared in were also examined. 

A support vector machine learning algorithm (SVM), decision tree (J48) and Naïve Bayes algorithm were used to develop predictive models to identify the gender of the politician featured in a given article.  The models were trained and evaluated using a 10-fold cross-validation approach. In this method, a dataset is randomly separated into 10 datasets of equal size and class distributions. For each of the folds, the classifiers learn from all but one of the 10 datasets and the model is tested on the remaining. Overall accuracy or cross validation is found by averaging the results of the tests on each of the 10 datasets. The best classifier is determined by the highest predictive accuracy.

The discriminative features that were identified by the most accurate machine learning algorithm were then analysed to access whether they constituted gender bias. This approach built upon a technique outlined by Diermeier et al. (2012) where discriminative features are ordered according to the weighting of each attribute assigned by the linear SVM (Chang and Lin, 2008; Guyon et al., 2002). The support vector machine learning algorithm ranks each feature according to its association with one category of texts. The highest ranking features are those that best discriminate between the texts. 

Machine learning algorithms assume an even split between classes. In most of the experiments in this research this even split naturally occurred. However, in the sub- corpora, in some experiments there was an imbalance in the data set. There are many approaches to dealing with such imbalance. The most effective approach in this research was under-sampling, which involved randomly sampling from the majority class, the same number of instances as in the minority class (Akbani et al., 2004). 

Features that distinguished between the coverage of male and female politicians were analysed based on researcher interpretation, incorporating related academic literature along with contextual information regarding political events. Concordance analysis of the features was used to assess the context in which they occurred in the text. Where numerous occurrences of the features appear in the text in differing contexts, these contexts are grouped based on a qualitative analysis of the content informed by existing literature on what constitutes gender bias, thus allowing recurring themes to be identified.

The time spent in office is a crucial factor contributing to the number of articles written about a politician. In this corpus, male politicians spent more time in office than female politicians. On an individual level also, politicians spent varying amount of years in office. In order to compare political coverage, equivalent timeframes were examined by scaling the coverage of each politician based on the time they spent in office.

\section{Analysis and Findings}

The findings of this research did uncover evidence of gender bias in the content of political news in Irish print media. However a potential positive bias was found in relation to the volume of coverage afforded to female politicians. This research also demonstrated how feature engineering could be used along with machine learning classification algorithms to uncover different aspects of gender bias in language. Unlike many applications of text classification that strive to attain the highest predictive accuracy, as Heyer et al. (2006) explained, in the context of media content analysis, lower accuracy can often yield the most significant results. Similarly, in this research while in quantitative terms, certain features uncovered relatively small-scale differences, these differences represented important manifestations of gender bias when combined with qualitative analysis. 

\subsection{Gender Bias in Volume of Media Coverage}

\begin{figure}
\centering
\includegraphics[width=0.50\textwidth]{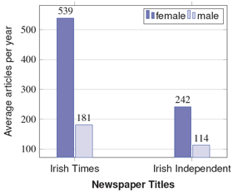}
\caption{Average Number of Articles Featuring Ministers .} \label{tbl_4}
\end{figure}

Female ministers were mentioned in Ireland’s leading newspapers 2.5 times as often as their male counterparts between 1997 and 2011 (Fig. \ref{tbl_4}). They appeared in headlines almost twice as often. This calculation considers the fact that in the time period that articles in this corpus were written, female ministers held ofﬁce for a total of 38.6 years and the comparable male ministers included in the research held ofﬁce for 84.1 years.  Given that previous research found the frequency of mentions of women relative to that of men can reﬂect gender bias (Bystrom et al., 2001; Fuertes-Olivera, 2007; Miller et al., 2010; Pearce, 2008), the results of this study could point towards a gender bias against male politicians.

However, studies of the quantity of coverage of politicians do not address the substance of the coverage. While female ministers may receive more than their male counterparts, this may not be an advantage if that coverage is more negative. Analysis of the differences in the content of articles featuring male and female politicians addressed this.

\section{Classification Accuracy: The Extent of Difference}
Classification accuracy results in this research show the extent of differences in coverage between male and female politicians. Classification accuracy gained using unigrams as features are shown in Table \ref{tbl_5} and Table \ref{tbl_6}. This analysis involved a broad range of experiments including different approaches to representing features (boolean, tf-idf, bag-of-words), and multiple machine learning algorithms in order to identify the best approach. These results show that the support vector machine learning algorithm was the most accurate in terms of identifying the gender of the minister featured in an article. Although the tf-idf representation yielded marginally higher accuracy than the Boolean representation, the Boolean representation, upon inspection, ranked features suggested gender bias highest. Neither removing stop-words, nor applying stemming, to unigrams improved the accuracy of the models nor the quality of the discriminative featured identified. 

\begin{table}[ht]
\centering
\begin{tabular}{c}
\includegraphics[width=0.66\textwidth]{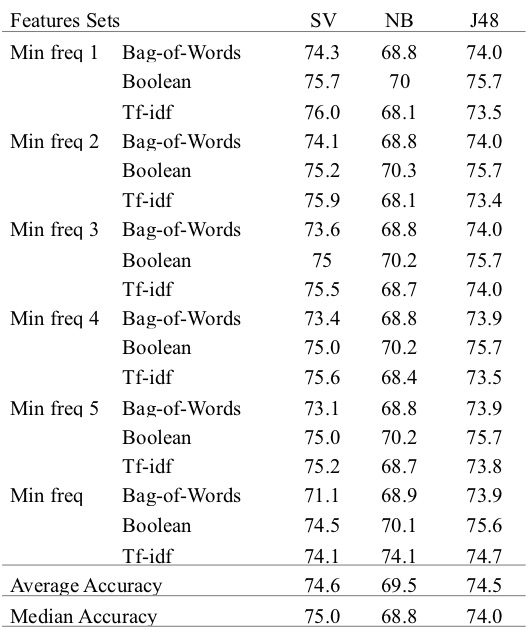}
\end{tabular}
\caption{Classifier Prediction Accuracy Using Unigrams as Features.} \label{tbl_5}
\end{table}

\begin{table}[ht]
\centering
\begin{tabular}{c}
\includegraphics[width=0.66\textwidth]{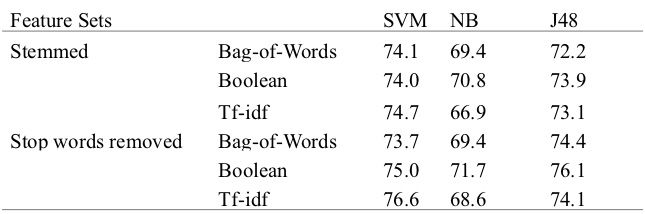}
\end{tabular}
\caption{Classifier Prediction Accuracy for Unigrams with Stop Words Removed and Stemming Applied.} \label{tbl_6}
\end{table}

The highest accuracy, or most difference in coverage, using adjectives as features was produced using a Boolean representation with the SVM classifier (Table \ref{tbl_7}). The discriminative features identified by this experiment were highly interpretable in relation to gender bias. For example, adjectives describing personal appearance associated with female ministers were notably absent, contrasting with previous research showing an excessive focus on female politicians’ appearance (Trimble et al., 2013). This demonstrated the benefits of using certain word types to target the analysis of the text.

\begin{table}[ht]
\centering
\begin{tabular}{c}
\includegraphics[width=0.80\textwidth]{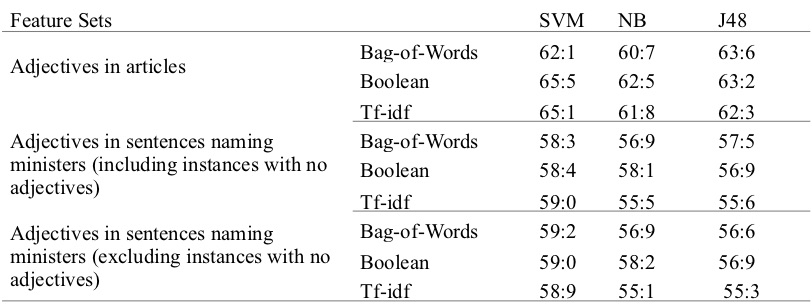}
\end{tabular}
\caption{Classifier Accuracies Using Adjectives as Features (baseline 52\%).} \label{tbl_7}
\end{table}

The predictive capacity of verbs in classifying texts according to the gender of the minister featured in the articles all remained close to 55 percent suggesting little variance in how verbs are used in articles featuring male or female politicians (Table \ref{tbl_8}). However, when verbs occurring only in sentences politicians were mentioned in were analysed, little gender difference was identified. The SVM classifier with a Boolean representation of features yielded the most accurate results.

Using the General Inquirer (Stone, 1966) lexicon of action and power words to extract features uncovered some differences in how power and active versus passive language is associated with male and female politicians (Table \ref{tbl_9}). Isolating words that co-occurred with sentences naming politicians rather than in the same article did not improve the predictive accuracy of the models. However the discriminative features that were identified using this approach were more closely related to the political subject.

\begin{table}[ht]
\centering
\begin{tabular}{c}
\includegraphics[width=0.66\textwidth]{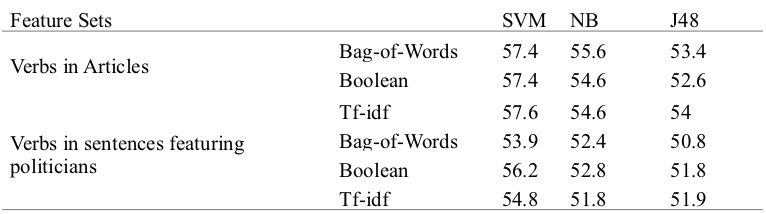}
\end{tabular}
\label{tbl_8}
\caption{Classifier Accuracies Using Verbs as Features.}
\end{table}

\begin{table}[ht]
\centering
\begin{tabular}{c}
\includegraphics[width=0.80\textwidth]{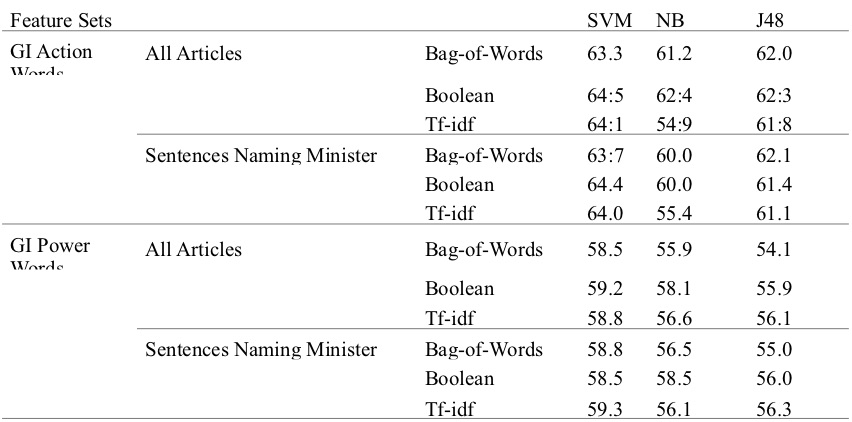}
\end{tabular}
\caption{Classifier Accuracies Using General Inquirer Lexicons as Features.} \label{tbl_9}
\end{table}

To explore whether gender bias is evident in the placement of articles in newspapers, section headings of newspapers were extracted and used as features for text classification yielding accuracies of almost 60 percent (Table \ref{tbl_10}). This pointed to some difference existing in the sections that male and female politicians appeared in and identified this as something to be qualitative investigated further.

\begin{table}[ht]
\centering
\begin{tabular}{c}
\includegraphics[width=0.50\textwidth]{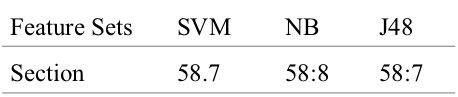}
\end{tabular}
\caption{Classifier Prediction Accuracy Using the Newspaper Section as Features.} \label{tbl_10}
\end{table}

While overall accuracies gained in machine learning experiments are an indicator of the strength of difference in coverage between male and female politicians, these differences could be attributed to multiple causes. It was necessary therefore to qualitatively inspect the differences uncovered. The next section presents this phase of the analysis.

\subsection{Discriminative Features: The Kind of Difference}

To uncover gender bias in the content of newspaper articles, the discriminative features identified by the SVM algorithm were analysed. This analysis was framed within the findings from previous research on the representation of female politicians in the media along with contextual knowledge of political events between 1997 and 2011. To verify the context of the words in the corpus, where necessary, concordance lines of the words were extracted from the corpus and qualitatively analysed to access whether the patterns identified by the classifier constituted gender bias. 

\subsection{Family Relationships and Roles}
Terms pertaining to family relationships and roles were associated with female ministers. For instance, the word ‘husband’ was more likely to be used in articles featuring a female politician in the headline (Table \ref{tbl_11}). However, the term referring to the spouse of male ministers, ‘wife’, did not appear as a discriminative feature in any of the classiﬁcation experiments. This imbalance in the references to spouses suggests a gender bias placing more importance is placed on the marital status of women (Baker, 2010; Fuertes-Olivera, 2007; Holmes, 1994). This reinforces previous studies showing that coverage of female politicians is more likely to focus on their personal family circumstances than male politicians (Brikse, 2004; Miller and Peake, 2013; Spears et al., 2000).

\begin{table}[ht]
\centering
\begin{tabular}{c}
\includegraphics[width=0.33\textwidth]{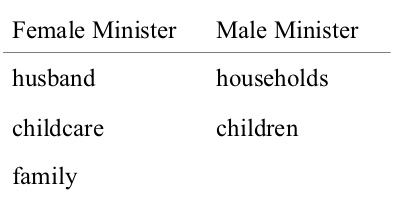}
\end {tabular}
\caption{Discriminative Features Associated with Family.} \label{tbl_11}
\end{table}

To further examine the context in which spouses of ministers were discussed, concordances of both words from the entire corpus were extracted. These were manually analysed to identify those instances that referred to a minister’s spouse. Analysis of concordance lines of the word ‘husband’ in the corpus showed that there were 48 mentions of the term referring to a minister’s spouse and 27 mentions of the word ‘wife’ referring to a minister’s spouse. Hence, for every year in ofﬁce, a female minister’s spouse is mentioned four times as often as that of a male minister’s (Table \ref{tbl_12}).

\begin{table}[ht]
\centering
\begin{tabular}{c}
\includegraphics[width=0.66\textwidth]{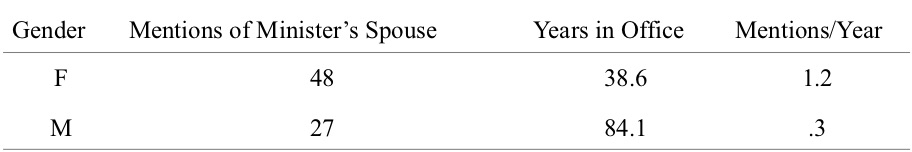}
\end {tabular}
\caption{Mentions of Spouse per Year in Office.} \label{tbl_12}
\end{table}

There were also differences in the context in which spouses were mentioned. This was examined by extracting concordance lines of words referring to the spouses of ministers. These were then grouped according to common themes. Analysis of the use of the terms ‘wife’ and ‘husband’ showed that the wives of minsters were often described as being ‘at home’ and were described as attractive or emotional. Husbands of ministers however were spoken of in discussions of family life and their relationships were portrayed as equal partnerships. Some examples how the words ‘wife’ and ’husband’ were used in relation to ministers are shown in Table \ref{tbl_13}.

\begin{table}[ht]
\centering
\begin{tabular}{c}
\includegraphics[width=0.90\textwidth]{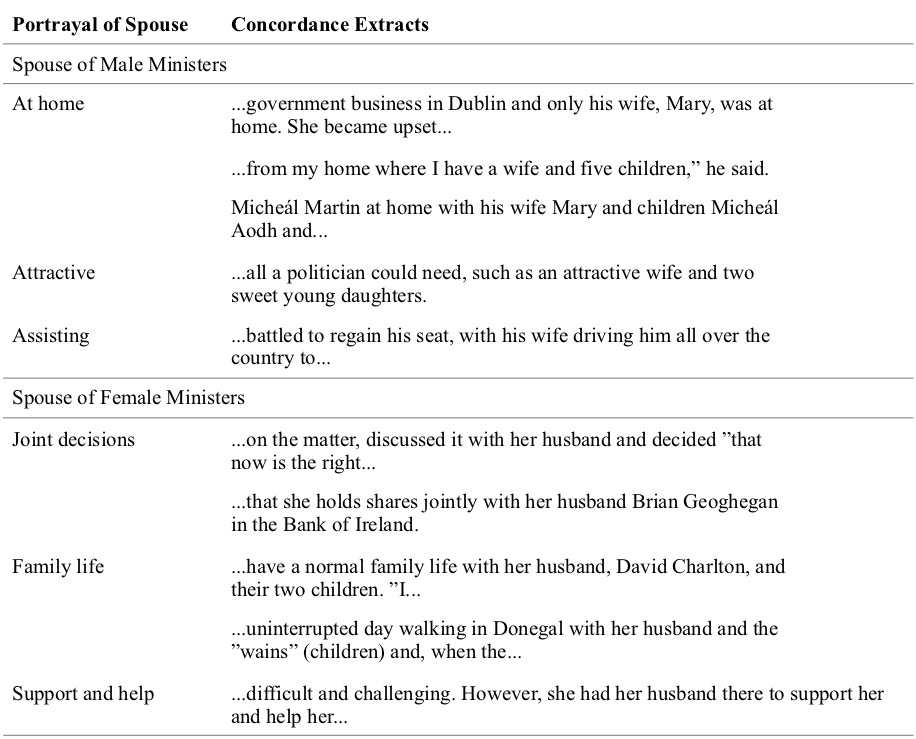}
\end {tabular}
\caption{Examples of Concordance Lines of Political Spouse.} \label{tbl_13}
\end{table}

The term ‘household’ is associated with male politicians. This term is commonly used in economic discourse referring to households as economic units. Concordance analysis showed that the term was used in the coverage of male politicians primarily to discuss public policy and economic issues pertaining to families. This contrasts with the association of female politicians with the word ‘family’. The term is more likely to be used in personal discussions of an individual’s family or in relation to social policy. References to ‘family’, taking into account the years spent in ofﬁce, were cited 2.5 times more often in articles featuring female ministers than in articles featuring male ministers. When only extracts that referenced a minister’s own family were analysed, the families of male ministers were mentioned more often per year in ofﬁce that the families of female ministers. This could be explained by the fact that a greater proportion of male ministers had children and were married. Each of these concordance extracts are shown in Table \ref{tbl_14} and are grouped according to a qualitative analysis of the themes that emerge from the excerpts. 

\begin{table}[ht]
\centering
\begin{tabular}{c}
\includegraphics[width=0.90\textwidth]{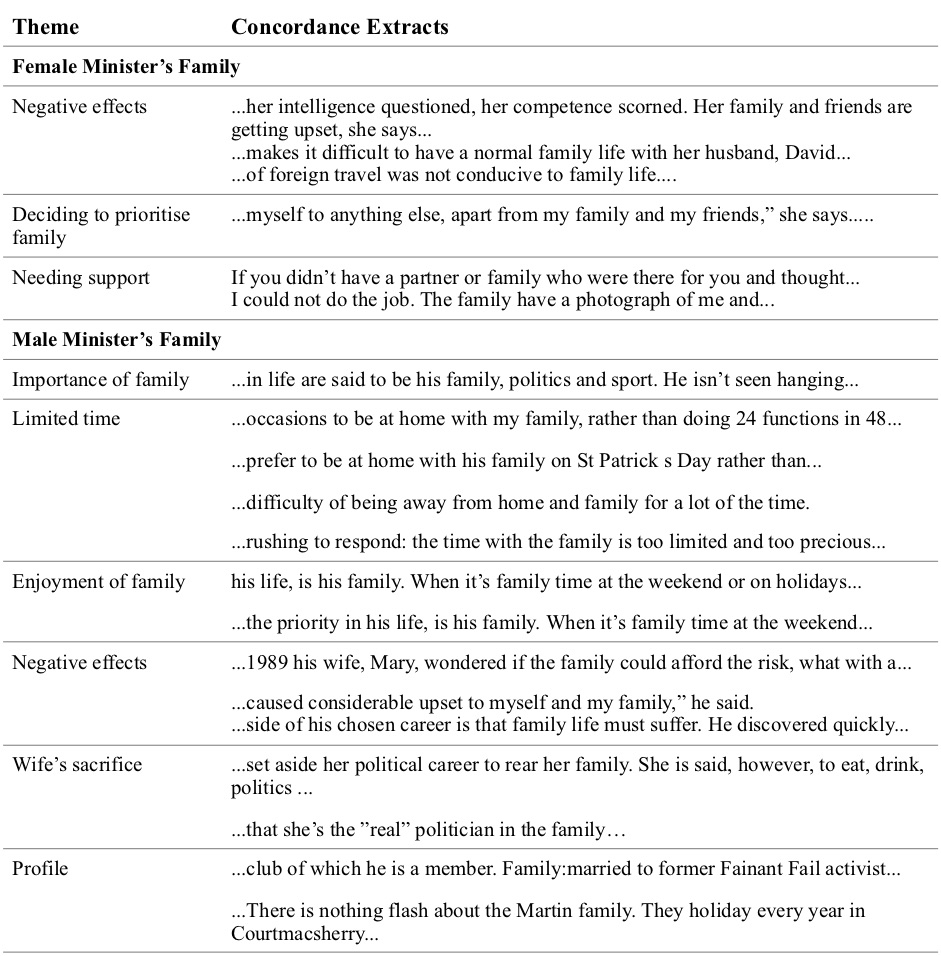}
\end {tabular}
\caption{Concordance Examples of Minister’s Immediate Family.} \label{tbl_14}
\end{table}

The family life of male ministers was portrayed in a more positive light than that of female ministers. Conﬂict emerging from the demands of political life and that of their family was a focus of coverage of both male and female ministers. This finding contradicts research that associates discussions of conﬂict between the demands of politics and family primarily with female politicians (Van Zoonen, 2006). However, the family life of the male ministers was also portrayed as a happy one despite the fact that they would have liked more time to devote to them. A similar positive portrayal of a female politician’s family life is lacking resulting in an overall focus on the negative aspects of combining politics with family life.

Concordance extracts of mentions of the word ‘family’ showed that 90 percent of the instances where policy was mentioned in terms of work-life balance and childcare issues, occurred in articles featuring female politicians. This corroborates findings from previous studies that uncovered a tendency in the media to associate female ministers with policy that is perceived as women’s issues regardless of their cabinet portfolio (Brikse, 2004; Carroll and Schreiber, 1997; Devitt, 1999; Kahn, 1994, 1996).

The word ‘children’ was identiﬁed as an important discriminative feature for articles featuring male politicians. Examination of concordance extracts of all 1,687 instances of the word children in the corpus showed that these primarily pertained to discussions of policy regarding health and social welfare.	As these issues most commonly concerned the Minister for Health, it was important to view the frequency of mentions factoring in the gender of the ministers. Factoring into account the time that male and female ministers spent as Ministers for Health and Children, the rates of mentions of the term are almost equivalent for male and female ministers.

The word ‘childcare’ was identiﬁed in the text classiﬁcation experiments as being associated with female ministers. A qualitative analysis of concordance lines showed that all references to ’childcare’ related to the need for an improved childcare system. Of the 83 mentions, only 12 of those were from male politicians. Hence for each year in ofﬁce, female ministers were associated with childcare 11 times more than their male counterparts, despite the fact that male politicians held portfolios related to this policy area for longer.

\subsection{Focus on gender}

Miller et al. (2010) and Norris (1997) cited the unnecessary focus on the gender of women in politics as evidence of gender bias in the media. The word ’female’ was identiﬁed in the classiﬁcation algorithms as being a discriminative factor identifying articles featuring female politicians. However, the word male is not associated with articles featuring male politicians. Analysis of the frequency of the words in the corpus showed that for each year in ofﬁce, the word ‘female’ was mentioned over 5 times as often in association with female ministers than the word ‘male’ was associated with male ministers.

Concordance analysis of the term ‘female’ found that references to the minister’s own gender accounted for 23 percent of these instances. In contrast, analysis of concordance extracts of the word ‘male’ showed that of all of the mentions of the word, not one referred to a male politician’s gender. This suggests gender bias in the form of an excessive focus on gender as outlined by Norris (1997) and Miller et al. (2010).
Gender equality policy was the issue under discussion in 20 percent of the occurrences of the word ‘female’. Men and women were also associated differently with these issues. Female ministers were associated with the issues more frequently and described as supporting them. However male minsters were only described as being under pressure to conform to them.

\subsection{Use of stereotypes}
An association was uncovered between a workplace culture linked to the consumption of alcohol and male ministers. This kind of political culture has been identiﬁed in Ireland as alienating women from politics (Joint Committee on Justice, Equality, Defence and Women‘s Rights, 2009). The word ‘drink’ was identiﬁed as the most discriminative verb associated with articles featuring male ministers. Analysis of concordance lines of the term showed that of all of the instances where a minister is portrayed as drinking alcohol, only one of those pertains to a female minister. All other references relate to a male minister’s personal consumption and purchase of alcohol.

In one example of a reference to a drinking culture, Mary Hanaﬁn, is described as being ‘pushed aside’ and a political appointment going instead to the Prime Minister’s ‘drinking buddy’. Such portrayals of a workplace culture may be a reﬂection of the political environment rather than a bias on the part of the media. However, systematic association of male politicians with an activity seen as a core part of the workplace culture of politics may serve to portray women’s involvement with politics as counter to gender stereotypes (Adcock, 2010).

The verb ‘beat’ was identiﬁed as the second most important verb associated with male ministers. Concordance analysis showed that these references were primarily mentioned within the context of sporting events and all except one of these were in association with male ministers. The exception featured a female minister leaving a sporting event early because she ‘looked bored’. This association of sporting events with male ministers while suggesting a stereotypical association of men with sport and women as being ’bored’ by it, could also be explained by the fact that there was a female minister for sport for only one of the years studied in this research.

Adcock (2010) outlined how sexualised identities are created in media portrayals of female politicians. The term ‘woo’ was highlighted as a discriminative feature associated with articles featuring female ministers. This term is commonly used in the context of an initiation of romance. Analysis of concordance lines of ‘woo’ show that all except one of the instances are used in articles featuring female politicians (Table \ref{tbl_15}).

The term was used metaphorically to describe how women encouraged people to adopt certain political stances or take certain actions. Describing female ministers metaphorically as ‘wooing’ other parties in negotiations while not using a word with similar connotations in connection with male politicians does suggest the creation of a sexualised identity for female ministers. The identity constructed is of active sexuality, contrary to ﬁndings from previous studies that found male sexuality to be portrayed in terms of predatory behaviour (Mills, 2002).

\begin{table}[ht]
\centering
\begin{tabular}{c}
\includegraphics[width=0.90\textwidth]{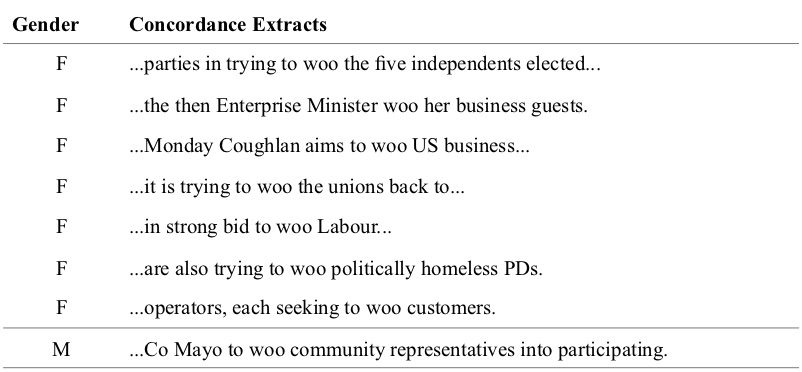}
\end{tabular}
\caption{Concordance Examples of the word ‘Woo’.} \label{tbl_15}
\end{table}

\subsection{Political style}
Differences in the portrayal of the political style and attitude of politicians in the media were identiﬁed by the classiﬁer using the General Inquirer Lexicon of action words as features. Of these, the most important discriminative feature for ministers were terms often used to describe an attitude to policy. The term ‘revoke’ was associated with male ministers while the term ‘embrace’ was associated with female ministers (Table \ref{tbl_16}).

Concordance analysis showed that the term, ‘embrace’, associated with female politicians, was used primarily as a metaphorical description of their attitude to public policy or political ideology. The word most associated with male politicians, ‘revoke’ was used in the context of opposing approaches to government policy of revoking legislation. These ﬁndings suggest what Adcock (2010) described as the portrayal of female politicians as ”condensing symbols” for the political ideologies they are afﬁliated with, referring to women being portrayed as being more unconditionally supportive of government policy. Men however, with the increased association with the term ‘revoke’, are portrayed as more confrontational. 

\begin{table}[ht]
\centering
\begin{tabular}{c}
\includegraphics[width=0.90\textwidth]{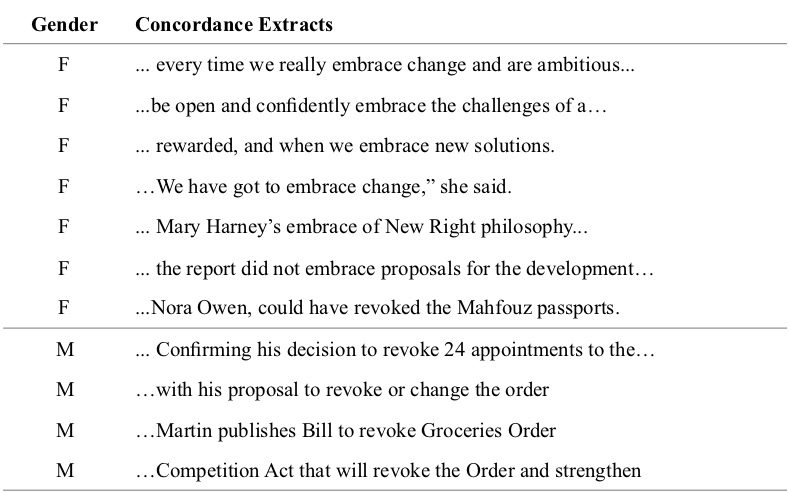}
\end{tabular}
\caption{Concordance Examples of Mentions of ‘Embrace’ and ‘Revoke’.} \label{tbl_16}
\end{table}

Other words from the General Inquirer lexicon of action that discriminated between male and female politicians referred to a politician’s communication style. The following are the discriminative terms in order of importance to each category:
Male Ministers: unite, cooperation, cultivate
Female Ministers: misrepresent, smear, discredit, ridicule, shout
Concordance analysis showed that there were very few occurrences of each word in the corpus. There were also differences in the sense in which the words were used. For example, uses of the word ‘smear’ showed that the all the female ministers were carrying out the action while the male ministers were the objects of them. 

The words were grouped according to whether they conveyed a positive or negative portrayal of communication style based on researcher interpretation. Of the words listed, the terms ‘unite’, ‘cooperation’ and ‘cultivate’ were judged as positive reﬂections of diplomatic communication style. The positive words directly correlated with those associated with male politicians (unite, cooperation, cultivate) and the negative words were those most associated with female ministers. Furthermore, while the words associated with female ministers were direct descriptions of speech acts, those associated with male ministers were descriptions of the outcomes or aims of political discourse.

The purpose of the speech acts associated with female ministers also differs from the political outcomes associated with male ministers. The aims of those associated with female ministers are to undermine an opponent while male ministers are associated with political diplomacy. This aligns with Gidengil and Everitt (2003), who found that female politicians are more likely to be described as having an aggressive political communication style.

\subsection{Personalised coverage}
Female ministers were referred to using their ﬁrst names more often than male ministers. This aligns with research which found that female politicians are more likely to be referred to in the media using informal versions of their names (Fox, 1997; Uscinski and Goren, 2011). At the text pre-processing stage of this research, names of politicians were replaced with gender-neutral terms that also documented the form of the name used. Results of text classiﬁcation experiments showed that naming using a minister’s ﬁrst name followed by their surname was associated with female ministers. Using surnames alone were more associated with male ministers. Given that higher status is associated with use of the more formal method of naming, using a surname only (Page, 2003), this ﬁnding suggests evidence of media bias.

A pattern of difference in how male and female politicians were associated with terms relating to power emerged in the use of the General Inquirer lexicon of power words as features. The top 5 power words associated with female politicians were ‘executive’, ‘distinguished’, ‘formidable’, ‘leader’ and ‘director’. The top features associated with male ministers were ‘jail’, ‘prohibition’, ‘squad’ and ‘chief’, suggesting power of institutions. However, concordance analysis showed that only two of these terms were used in a sense that pertained to the ministers themselves and issues of power. For example, the term ‘distinguished’ was used primarily as a verb that had no relation to portrayals of powerful people. Squad was used metaphorically to refer to football teams. This demonstrates the importance of veriﬁcation of these patterns through qualitative analysis of the context of word use.

Baker (2010) noted that only women were described as ’formidable’ in analysis of a corpus of modern British English. This term was also associated with women in this research. Concordance analysis showed that the term ’formidable’ was used to portray both male and female politicians. However, accounting for the time the ministers spent in office,  ‘formidable’ is used to describe female ministers almost four times as often as male ministers. These ﬁndings suggest that in the same way that research has examined descriptions of female politicians as strident (Childs, 2004), further research would be useful to assess the implications of describing them as ‘formidable’.

\subsection{Masculine narrative/metaphor}

Media coverage of politics has been critiqued for using metaphors of war and sport traditionally associated with men in political discourse (Gidengil and Everitt, 1999; Trimble et al., 2013). While terms pertaining to activities of war were not identiﬁed by the classiﬁers as terms associated with ministers of either gender, terms pertaining to power inherent in institutions of the state, including the military, and positions of power within it are associated with male ministers. Terms associated with competitive sport are also linked with male ministers. There were also more terms associated with male ministers then female ministers. This corroborates the ﬁnding from Koller’s (2004) that there are more gender schemas evident in business magazines for businessmen than for businesswomen reﬂecting the fact that ”discursive and cognitive structures determined by hegemonic masculinity show more cultural models of masculinity than of femininity”. The following are the discriminative features which pertain to governmental power and sport:
\begin{itemize}
\item\textbf{Male ministers:} jail, prohibition, squad, chief, politician, chairmen, council, mayor, president, championship, aristocracy, champion, guard, commander, presidency\\
\item\textbf{Female ministers:} government, ofﬁcer
\end{itemize}

Analysis of the concordance lines of these terms showed that the words not frequently used in direct reference to the ministers themselves. However, the systematic association of male politicians with newspaper articles mentioning high-status roles within the state, military and competitive sport does suggest a bias that excludes female ministers from such discourse. Female ministers are associated with two terms that are linked to institutions of the state, ‘government’ and ‘ofﬁcer’. The relative absence of association of female politicians with the police force, army or high-status institutional roles suggests that female ministers may be disassociated from those positions and institutions of government. Further research is needed to attribute to bias on the part of the newspaper to this.

The link in newspaper coverage between male ministers and terms pertaining to the police force and the army suggest that male politicians in Ireland are associated with ”military masculinity” (Higgins, 2010, p.151). The pattern identiﬁed by the classiﬁcation experiments associating male ministers with power linked to the military and institutions of the state can be interpreted as perpetuating gendered stereotypes regarding women and their relationship with power (Okimoto and Brescoll, 2010).

The terms ‘riot’ and ‘ban’, which are reminiscent of military actions, were also associated with male politicians when the General Inquirer Lexicon of political words was used to extract features. However, in contrast to these ﬁndings, female politicians are associated with the term ‘invasion’, which can be used in the context of an act of war while male politicians are associated with words more associated with peace negotiations including ‘conciliation’, ‘mediation’ and ‘neutrality’. Concordance analysis of these terms however showed a wide variety of contexts in which the words were used. This illustrates the importance of verifying quantitative results with qualitative analysis.

\subsection{Newspaper section}
When newspaper section headings were used as features the models the machine learning algorithms generated yielded contradictory results. While there was evidence of stereotypical associations of male and female ministers with newspaper sections, female ministers also seem to dominate the higher proﬁle sections of the newspapers. Similar to the positive bias suggested by the volume of coverage afforded to female ministers, these results suggest a bias in favour of female politicians.

Male politicians feature more in sections stereotypically associated with male pursuits such as motoring, farming and sport. They are also more prevalent in the sections relating to science and international politics. Female ministers appear most in the letters, opinion, entertainment and lifestyle sections. This aligns with research showing more personalised coverage of female politicians in the media then their male counterparts (Devitt, 1999; Garcia-Blanco and Wahl-Jorgensen, 2011; Ross and Sreberny, 2000). Female politicians also featured more in the health section despite male ministers holding the office of minister for health for longer than female ministers. This suggests a gender bias identiﬁed by Brikse (2004) and Carroll and Schreiber (1997) where female politicians are associated with health and social issues regardless of their actual role. However, female ministers also appear in the politics section, feature articles, editorial pages and front pages of newspapers more than male ministers, which points towards gender bias in their favour.

\section{Conclusion}
The findings of this research showed that coverage of politicians in Irish newspapers contains evidence of gender bias. The research also demonstrated how text classification could be used to analyse texts for evidence of gender bias. Machine learning algorithms highlighted differences in how Irish newspapers covered male and female politicians and analysis of these differences showed that many could be attributed to gender bias. 

There were two key objectives of this research. The first objective was to propose and test a new methodological approach to analysing gender bias in text. The purpose of this was to contribute to the existing range of methods available to researchers using a corpus approach in analysing gender. The second objective was to examine whether there is bias in the coverage of politicians in Ireland and what the nature of it is. Ireland currently holds the 92nd position globally for the level of women’s participation in politics (IPU, 2013). Given that gender bias in the media has been shown to discourage women from entering politics (Fox and Lawless, 2004; Heldman et al., 2005), it is important to examine media coverage of politics for evidence of gender bias. To date there has been little analysis of bias in the Irish media (Ahmad et al., 2011; Brandenburg, 2005) and this research has addressed this issue.

The ﬁndings showed that there was evidence of gender bias in the coverage of female politicians in Ireland. In previous studies of gender bias in the media, bias tends to be analysed in terms of the quantity and the content of the coverage (Gidengil and Everitt, 1999). The findings of this research showed a distinct positive bias towards female politicians in terms of the volume of coverage they received and their placement within newspaper sections. However, differences were evident in how male and female politicians were featured in articles that may be attributable to gender bias. 

Differences in coverage included an increased focus on family relationships and roles, differences in how they were associated with policy, an excessive focus on the gender of female politicians, more negative portrayals of political style and more personalised coverage. This research highlighted the ways in which gender bias is expressed in the coverage of politicians in Irish newspapers. The kind of bias identified aligned with bias found in some related studies. However, unexpected forms of gender bias were also uncovered demonstrating the value of exploring data-driven approaches to text analysis enabled by machine learning.

%
%
%
%

\end{document}